\documentclass[runningheads]{llncs}

\usepackage{amsmath}
\usepackage[T1]{fontenc}
\usepackage{graphicx}
\usepackage{soul}
\usepackage{wrapfig}
\usepackage{url}
\usepackage[utf8]{inputenc}
\usepackage[small]{caption}
\usepackage{booktabs}
\usepackage{algorithm}
\usepackage{algorithmic}
\usepackage[switch]{lineno}
\usepackage{comment}

\usepackage{nicefrac}
\usepackage{caption}
\usepackage{subcaption}
\usepackage{url}

\usepackage{hyperref}
\usepackage[textsize=tiny]{todonotes}

\usepackage{color}

\begin{document}
\title{Efficiently Training Neural Networks for Imperfect Information Games \\ by Sampling Information Sets}
\titlerunning{Efficiently Training Neural Networks}

\author{Timo Bertram\inst{1}
\and
Johannes F\"urnkranz \inst{1,2}
\and
Martin M\"uller \inst{3}
}
\authorrunning{T. Bertram et al.}
%
\institute{Johannes Kepler University, Linz, Austria \and
LIT AI Lab, Linz, Austria \and
University of Alberta, Edmonton, Canada\\
\email{\{tbertram,juffi\}@faw.jku.at}\\ 
\email{mmueller@ualberta.ca}}
\maketitle              
\begin{abstract}


In imperfect information games, the evaluation of a game state not only depends on the observable world but also relies on hidden parts of the environment. As accessing the obstructed information trivialises state evaluations, one approach to tackle such problems is to estimate the value of the imperfect state as a combination of all states in the information set, i.e., all possible states that are consistent with the current imperfect information. In this work, the goal is to learn a function that maps from the imperfect game information state to its expected value. However, constructing a perfect training set, i.e. an enumeration of the whole information set for numerous imperfect states, is often infeasible. To compute the expected values for an imperfect information game like \textit{Reconnaissance Blind Chess}, one would need to evaluate thousands of chess positions just to obtain the training target for a single state. Still, the expected value of a state can already be approximated with appropriate accuracy from a much smaller set of evaluations. Thus, in this paper, we empirically investigate how a budget of perfect information game evaluations should be distributed among training samples to maximise the return. Our results show that sampling a small number of states, in our experiments roughly 3, for a larger number of separate positions is preferable over repeatedly sampling a smaller quantity of states. Thus, we find that in our case, the quantity of different samples seems to be more important than higher target quality.

\end{abstract}
\section{Introduction}

Imperfect information games, which are characterised by unobservable aspects, are an important part of Game AI research. In recent years, they have received increased attention due to the inherent complexity of managing uncertainty. This category encompasses a wide array of games, spanning from classical card games like Poker and Bridge to adaptions of traditional board games such as Dark Hex and Reconnaissance Blind Chess and real-time video games like Starcraft, Dota II and Counter-Strike. Thus, we see much interest — commercially and scientifically — in mastering this category of games. However, methods that conquered many classical perfect information games, e.g. AlphaZero \cite{silver2018general}, do not easily carry over to imperfect information games \cite{PlayerOfGames}. In imperfect information settings, decisions are typically based on a fusion of the public information and an implicitly learned or directly computed expected value of the hidden information. While there are several approaches to learning evaluations implicitly (see Section~\ref{sec:Related_Work}), we here focus on learning them explicitly in a supervised fashion.

At every decision point of an imperfect information game, the set of all possible states from one observer's perspective is called an \textit{information set}. For sequential games such as \textit{Poker} and \textit{Reconnaissance Blind Chess,} it is possible to enumerate the states in the information set, thus allowing for approximation of the total evaluation of the imperfect information state as an aggregation of the single evaluations in the set. However, a perfect enumeration of the set can be vastly too costly. Thus, we aim to learn an evaluator which encapsulates the relationship between an imperfect information state and its aggregated target evaluation, enabling fast evaluation of states. In training, however, the same problem persists; generating perfect target values for states requires an unreasonable amount of evaluations. To reduce this load, we only use parts of the information set for target generation and investigate how many individual states are required for training.

\begin{figure*}[h]
    \centering
    \includegraphics[width = \linewidth]{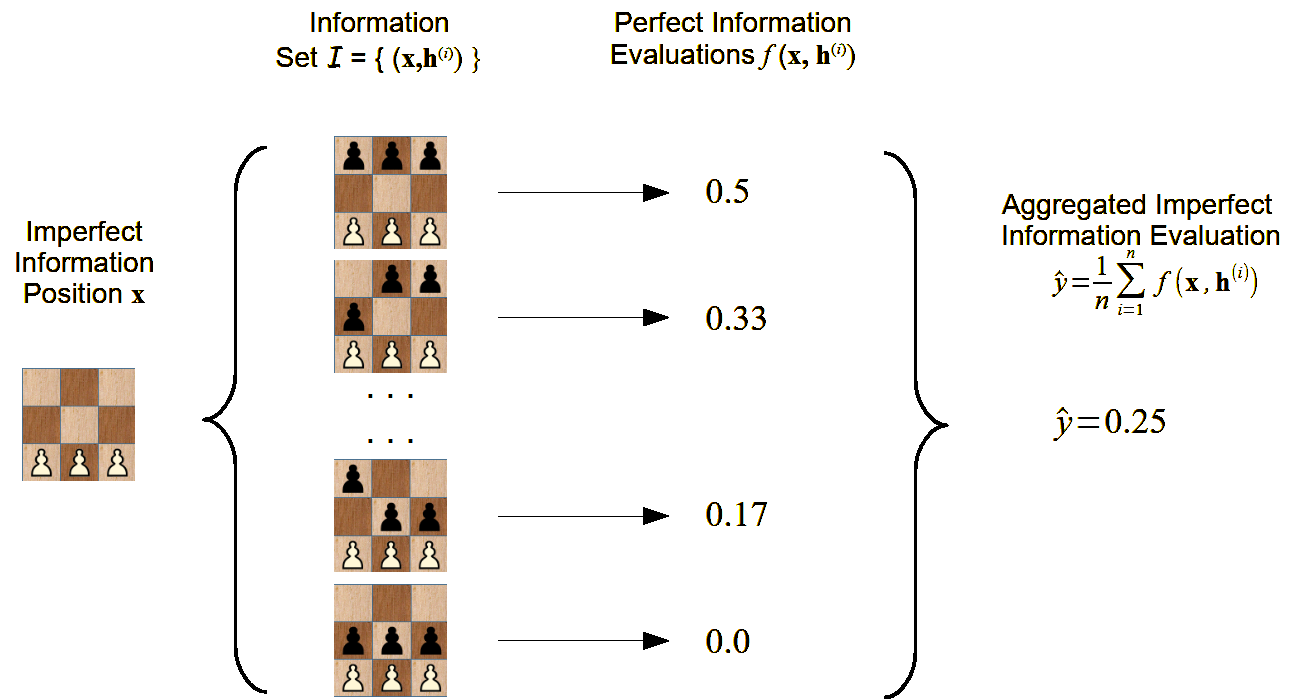}
    \caption{Estimating the value of an imperfect information position (left) as the average of the perfect information evaluations of all positions in the information set.}
    \label{fig:evaluation}
\end{figure*}

This setting is motivated by work in \textit{Reconnaissance Blind Chess} (RBC), a chess variant where parts of the board are not observable, and thus agents have to evaluate imperfect information game states. Many leading programs approximate these values by evaluating the states in the information set with a conventional perfect information chess engine and averaging the results. While these averages can serve as training signals for a model, evaluating every single board state in each information set is unreasonable. In this work, we question how we can best invest a fixed budget of total game state evaluations; should they be spread out to obtain inaccurate estimates for more training data or should they provide more reliable training signals for fewer data points? We begin this work by outlining the problem more formally (Section~\ref{sec:problem_statement}) and give a brief overview of related work and problems (Section~\ref{sec:Related_Work}). 
Subsequently, we empirically investigate the problem in two different imperfect information games:
\begin{itemize}
    \item \textbf{Heads-Up Poker}: Here we evaluate a 2-card hand without knowledge about the community cards or the opponent's cards, sampling from information sets of possible completions to estimate the win rate of a hand.
    \item \textbf{Reconnaissance Blind Chess (RBC)} \cite{RBC-tournament-1}: In this chess variant, the opponent's moves are often uncertain because of limited information. We receive an estimated evaluation of the imperfect information game state based on possible perfect game evaluations obtained from a conventional chess engine.
\end{itemize}

We summarise our results in Section~\ref{sec:summary_conclusion} and give an outlook on potential future extensions in Section~\ref{sec:future_work}.

\section{Problem Statement}
\label{sec:problem_statement}
 
We formalise the problem as follows: Given is a dataset of examples $\mathcal{D}  = (\mathbf{x}_i,y_i) \subset X \times Y$, where each label $y_i = f(\mathbf{x}_i,\mathbf{h}_i)$ is determined by a function $f$, dependent not only on the observable information $\mathbf{x}_i$, but also on the hidden information $\mathbf{h}_i$. Our goal is to find a function $g(\mathbf{x})$ which approximates $f(\mathbf{x},\mathbf{h})$, such that $\forall i \in \{1,..,|D|\}: g(\mathbf{x}_i) \approx f(\mathbf{x}_i, \mathbf{h}_i)$. This task is non-trivial, and such a function $g$ does not generally exist, as the same observable $\tilde{\mathbf{x}}$ can occur multiple times with different labels because, in general, $f(\tilde{\mathbf{x}},\mathbf{h^{(1)}}) \neq f(\tilde{\mathbf{x}},\mathbf{h^{(2)}})$ for $\mathbf{h^{(1)}} \neq \mathbf{h^{(2)}}$.

Our motivation for this problem originates from imperfect information games, where the information set represents all possible game states given one player's information. In several such games, remarkable performance has been achieved by basing the imperfect information gameplay, whether implicitly or explicitly, on perfect information evaluations of states in an information set \cite{bluml2023alphaze,bertram2022supervised,browne2012survey}. For example, the value of a player's hand in \textit{Poker} can be estimated as the expected value of the hand over all possible variations of the community and opponent's cards. 
Similarly, many strong  \textit{RBC} agents rely heavily on chess engines for evaluating conventional chess positions \cite{RBC-tournament-1,RBC-tournament-2,gardner2022machine} and approximate the imperfect information state with the expected values of the states in the information set. Figure~\ref{fig:evaluation} illustrates this situation with a schematic $3 \times 3$ chess board where only the white pieces are public and the evaluation of this board is obtained by averaging the (hypothetical) evaluations of all possible configurations of black pieces.

It is important to acknowledge the limitations of basing imperfect information policy fully on perfect information evaluations, as it is trivial to construct counterexamples where this fails. Despite its fundamental problems, the idea of estimating the evaluations by sampling from information sets has been successful in many algorithms. For example, using perfect information search on sampled states has been used in games such as Bridge \cite{GIB}, Skat \cite{Long2010} and Scrabble \cite{lig*Sheppard99}. In particular, Bl\"uml et al. \cite{bluml2023alphaze} have noticed that perfect information evaluations for imperfect information games perform surprisingly strongly. On a smaller scale, one can also relate this concept to the central idea of Monte-Carlo Tree Search. There, nodes in the game tree are evaluated based on stochastic rollouts, which results in a noisy but still useful signal. Similarly to the problem at hand, one can use more sophisticated, but also slower, rollout strategies to receive more accurate signals or more frequent fast and noisy rollouts.

\begin{figure*}[t]
    \centering
    \includegraphics[width = \linewidth]{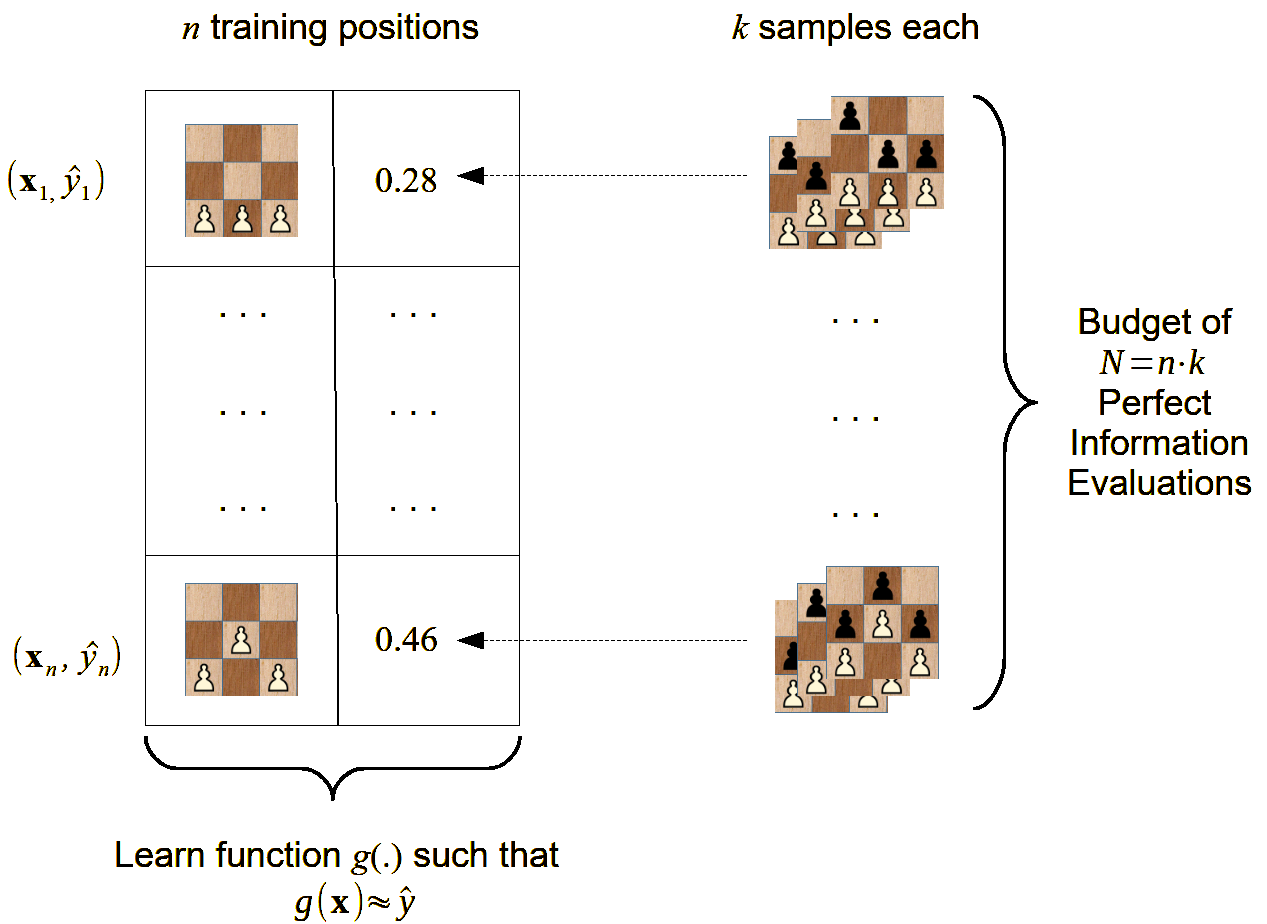}
    \caption{Learning an imperfect information evaluation function from $n$ examples, for which the target evaluation is estimated from $k$ position, using a constant budget of $N = n \cdot k$ perfect information evaluations.}
    \label{fig:learning}
\end{figure*}

\clearpage
The central objective of this work is to learn the function $g$ which receives the public information of a state $\mathbf{x}$ and approximates the expected value of that state, without the need for its (potentially expensive) explicit computation by iterating over the information set:

\begin{equation}
g(\mathbf{x}) \approx \hat{y} = \sum_\mathbf{h} P(\mathbf{h}|\mathbf{x}) \cdot f(\mathbf{x},\mathbf{h}) 
\label{eq:expected_f}
\end{equation}
\noindent
Here, $\mathbf{h} \in \mathcal{I}$ are all possible configurations of private information that are part of the information set $\mathcal{I}$, $f$ is an evaluator of a perfect information state and $P$ is a function which gives the probability of each hidden state for the given configuration $\mathbf{x}$. In practice, $P$ can be influenced by, among others, stochastic environments or adversary's (hidden) policies. In all cases, it is assumed to be an unobservable and unalterable part of the domain. In our experiments, as in Figure~\ref{fig:evaluation}, we assume that all possible determinations are equally likely, i.e. $P(\mathbf{h}|\mathbf{x}) =\nicefrac{1}{|\mathcal{I}(\mathbf{x})|}$ but one could also directly learn meaningful weights for the positions in the information set from past behaviour or observations \cite{bertram2023weighting}.

A simple strategy to learn $g$ is to collect samples of the form $(\mathbf{x}_i,\hat{y}_i)$, i.e., to compute the exact value $\hat{y}_i$ as in Equation \eqref{eq:expected_f} for many training positions $\mathbf{x}_i$, and to use supervised learning to learn the function $\hat{y}_i = g(\mathbf{x}_i)$ from these samples. However, this approach generally is too costly due to the potentially large size of information sets, so obtaining a single $\hat{y}_i$ can require thousands of evaluations. Alternatively, $\hat{y}$ can be approximated by randomly sampling only a few of the possible $\mathbf{h}^{(j)}$, resulting in less accurate training signals $\dot{y}$ at a lower computational cost. 

In our work, we aim to answer a fundamental question: Given a fixed budget of $N$ perfect information evaluations, how should we generate the training data for the learner? Options range from generating $N$ different training examples $\mathbf{x}_i$, each evaluated with one random sample, over using a fixed number of $k$ evaluations to generate targets for $n = \nicefrac{N}{k}$ positions, up to exhausting the budget with exactly computing $\hat{y}$ for as many examples as possible. This trade-off between the training set size $n$ (the number of distinct $\mathbf{x}_i$) and label quality (the number of evaluations $k$ used to estimate the intended target values $\hat{y}_i$ for each $\mathbf{x}_i$) forms the core focus of this paper.

\section{Related Work}
\label{sec:Related_Work}
The problem formulated in Section~\ref{sec:problem_statement} is multifaceted and occurs in several learning paradigms, thus we can only give a brief overview of how it manifests in practice.

Several learning settings may be viewed as special cases of this formulation. Conventional supervised learning emerges when $\mathbf{h}_i = \emptyset, \forall i$, i.e. when no hidden information determines $y_i$. Similarly, learning from noisy labels can be formulated with a single hidden variable $\mathbf{h}_i$, which determines whether the original label remains intact or is corrupted. Knowledge of this hidden information makes the underlying function $f$ deterministic, but $g$ does not have access to the information about the corruption.

Our setting is also closely related to research in the area of noisy labels \cite{snow2008cheap,khetan2018learning}, which extends to crowdsourcing \cite{sheng2008get,karger2014budget} and aggregating labels from different labellers. It also shares commonalities with active learning \cite{settles2012active}, which in our case is a problem of deciding whether to re-sample an existing example to improve label quality or to obtain a new sample to increase overall training data quantity. Importantly, most of this research aims to improve data distribution to the labellers or to reduce bias post-sampling, which differs significantly from choosing a sampling frequency a priori. In addition, noisy classification differs substantially from noisy regression.

In the context of imperfect information games, numerous approaches exist to, explicitly or implicitly, evaluate an information set. Techniques such as Perfect Information Monte Carlo \cite{long2010understanding,furtak2013recursive} combine evaluations of different perfect information searches into a policy for the imperfect information state and Information Set Monte Carlo Tree Search \cite{6031993} operates on information sets. Counterfactual Regret Minimization \cite{zinkevich2007regret}, as well as its successors and ReBeL \cite{brown2020combining} learn the utility of individual information sets through self-play. Recent work by Bl{\"u}ml et al.\cite{bluml2023alphaze} samples individual world states and constructs imperfect information policies based on their evaluations. In essence, most techniques for solving imperfect information games involve estimating the value of information sets, further motivating the importance of the question which we aim to answer.

Finally, the general idea that sampling more states from information sets will lead to a more accurate estimate of its overall value is simply an instance of the law of large numbers and thus finds application in a variety of problems. This trade-off between quantity and quality of evaluations is analogous to the choice of rollout policy in classical Monte Carlo Tree Search \cite{browne2012survey}, where one has to decide between random rollouts (fast, and thus allowing a larger quantity, but less informative) and more sophisticated rollout-policies (slower, thus limiting their number, but better at approximating true behaviour). However, one has to consider that MCTS differs slightly because UCB will over time re-visit promising states, thus allowing the resampling of important ones.

\section{Experiments}
\label{sec:experiments}

Here, we present a series of experiments designed to investigate the trade-off between target quality and training data quantity. The learner has a limited budget of total evaluations $N$ and can decide how many evaluations $k$ should be spent on each training example. Each evaluation yields one value, which are aggregated into a training target by averaging.

\subsection{Texas Hold'em Poker}
\label{sec:Poker}

The first experiment aims at a real-world setting where we have to balance the label accuracy with the number of total training examples seen. Here, the learner aims to estimate the win probability of a given 2-card hand of cards in two-player heads-up poker. This is a direct implementation of the problem outlined in Section~\ref{sec:problem_statement}. Pre-computed win-rate tables for any given hand exist, so one should regard this as a proof-of-concept rather than a useful application.

\begin{figure}[h]
    \centering
    \includegraphics[width = 0.8\linewidth]{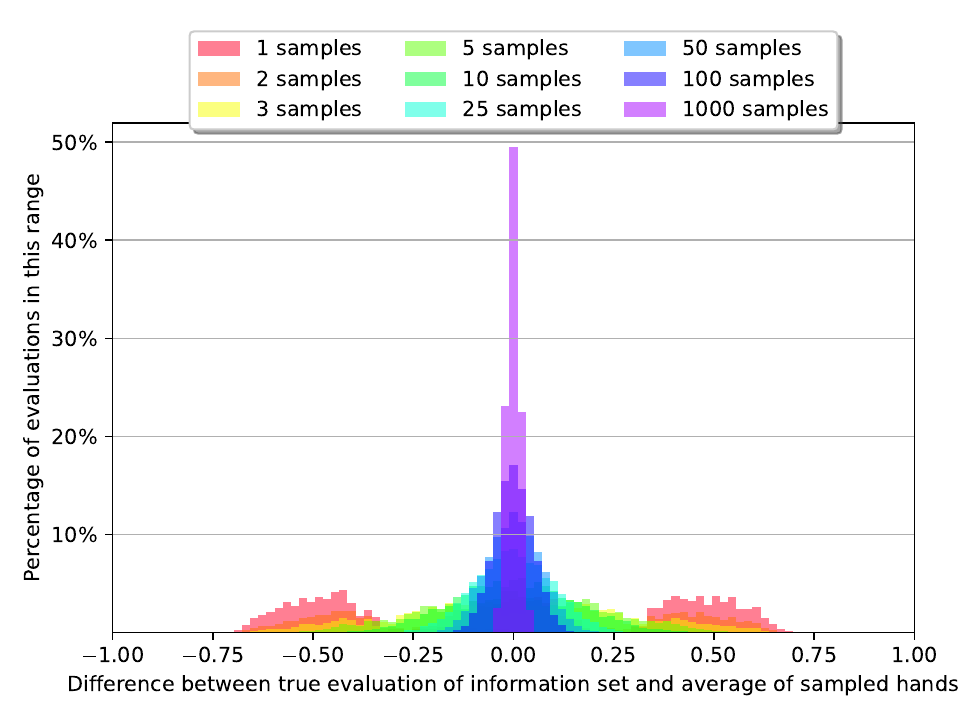}
    \caption{Monte-Carlo estimate of error in evaluating the change of winning with a given hand pre-flop- in heads-up poker. Estimations are computed by averaging over n samples of possible opponent hands and rivers.}
  \label{fig:poker_eval_error_samples}
\end{figure}

\subsubsection{Setup}
In principle, for a given hand $\mathbf{x}$, $g(\mathbf{x})$ could be directly computed as the average over all possible hidden contexts $\mathbf{h}_\mathbf{x}$, but doing so requires a large amount of computation. Without accounting for symmetry, a player can have $\binom{52}{2} = 1326$ unique Poker hands. One would need to compute all possible arrangements of the remaining cards into two opponent cards and five community cards, i.e. $\binom{52}{2} \cdot \binom{50}{2} \cdot \binom{48}{5}= 2{,}781{,}381{,}002{,}400$ total combinations. For each of these combinations, one needs to evaluate which player won the game and average this for all configurations that pertain to the same player's hand to estimate the overall winning probability of that hand. While public data for the win-chances of a hand exists, such data is only available for the most popular games and computing them is much costlier in other games with higher degrees of uncertainty or more expensive state evaluations. Thus, we aim to decrease the computational cost by only sampling parts of the information set instead of enumerating it entirely, and thus allow for the extension of the concept to a larger variety of applications.
    
When training the learner, we sample $k$ different configurations of cards for the given hand, evaluate the result of each configuration ($0$, $1$ or $0.5$), and train the network to predict the mean of all $k$ samples. Sampling more combinations leads to a smaller difference between the estimate and the ground truth, but results in a smaller amount of total hands seen when equating by the total number of evaluations.

\subsubsection{Results}

As a first estimate, Figure~\ref{fig:poker_eval_error_samples} shows the discrepancy between an estimated hand strength through evaluations and the mathematical true win chance. Notable, with only a single sampled configuration, it is impossible to exactly receive the true win chance of most hands as the only possible results are $0$, $0.5$, and $1$, thus resulting in three error clusters for one sample.

\begin{figure*}
    \centering
    \includegraphics[width = \linewidth]{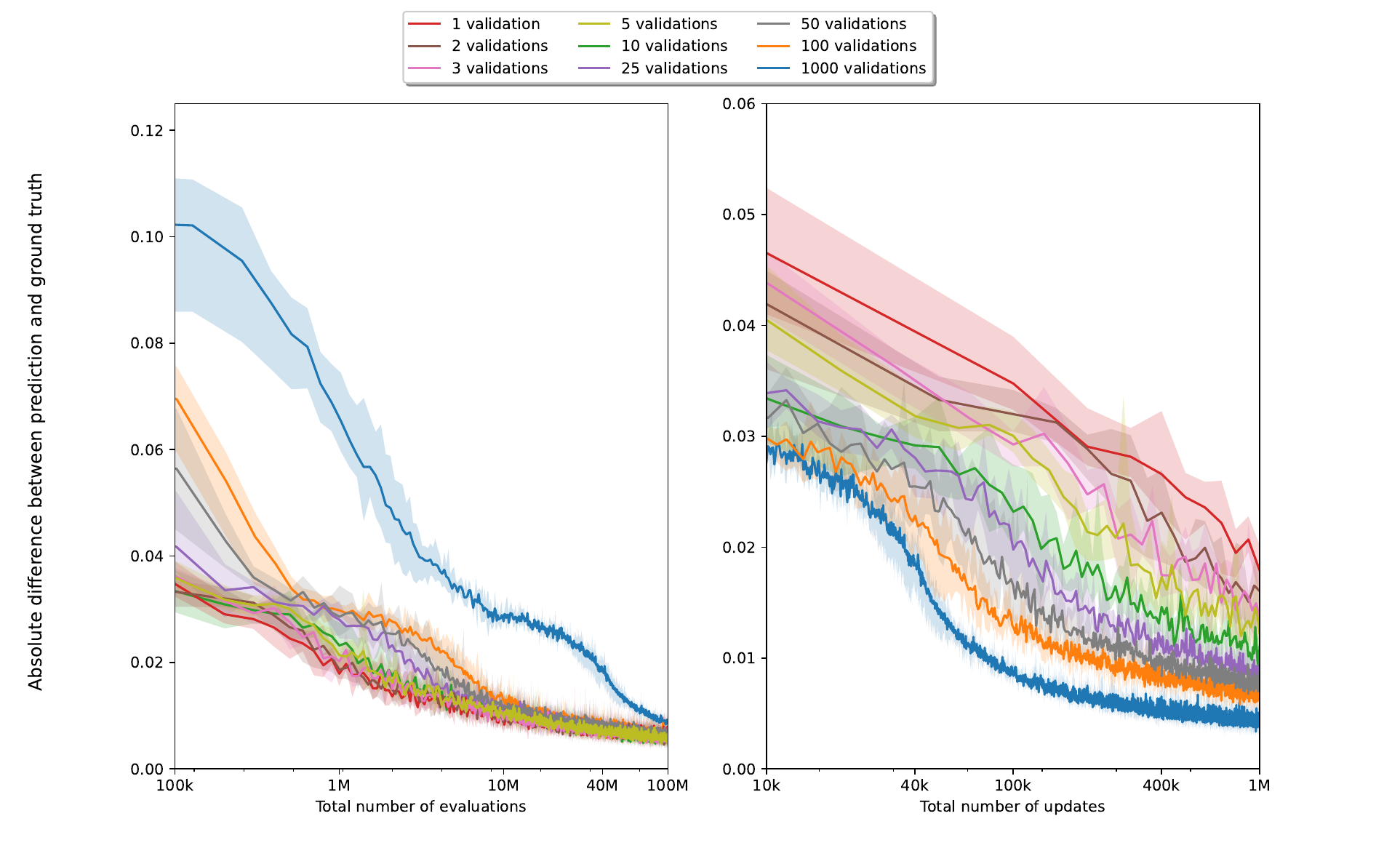}
    \caption{Average training curves of learning to evaluate a poker hand with different numbers of evaluations per training example. The $x$-axis is logarithmically scaled either by the total number of hand evaluations (top) or by the total number of update steps made (bottom). }
    \label{fig:poker_training_curve}
\end{figure*}

\begin{figure}[h]
    \centering
    \includegraphics[width = \linewidth]{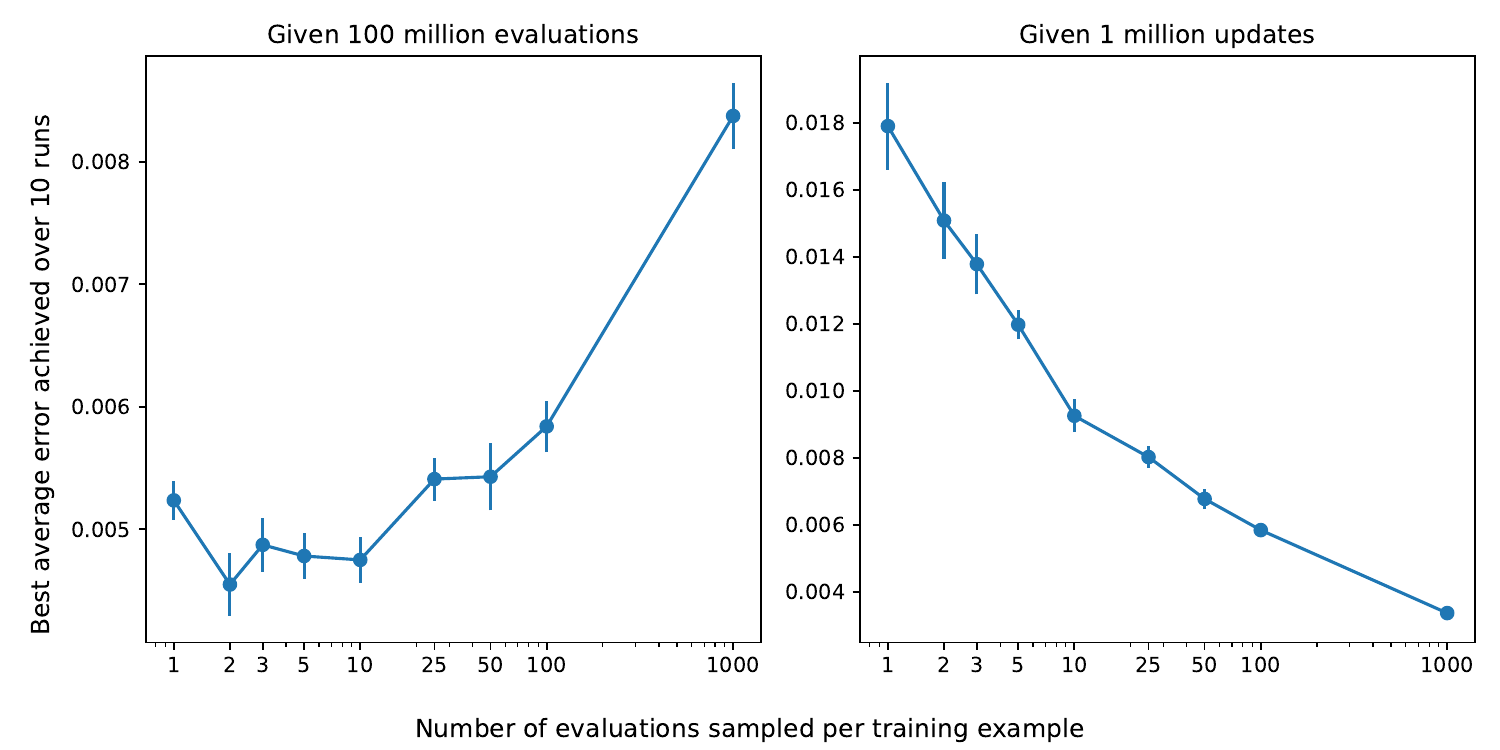}
    \caption{Average lowest received error for the different options of hand validations, given a total budget of either 100M hand evaluations or 1M training updates.}
    \label{fig:poker_best_version}
\end{figure}

The training process (Figure~\ref{fig:poker_training_curve}) shows that training with fewer evaluations per sample leads to much quicker progress when regarding the performance in relation to the total number of evaluations, but when the examples have higher-quality evaluations, each update is more meaningful. When comparing the best versions (Figure \ref{fig:poker_best_version}), we see that even when equating for the total number of evaluations, using a single evaluation leads to worse peak results than using two, three, five, and ten sampled evaluations. As it should be, equating training updates makes more evaluations perform strictly better than less. All in all, these results suggest that multiple samples are useful for this setting, but naturally, spending too much computation on a single example degrades the overall performance as the total training size diminishes.

We extend our experiments in the next section, where we regard a similar setting. For that, pre-computed evaluations do not exist and approximating through sampling is a realistic setting.

\subsection{Reconnaissance Blind Chess}
\label{sec:RBC}

\textit{Reconnaissance Blind Chess}.\footnote{\url{https://rbc.jhuapl.edu/}} \textit{RBC} is an imperfect information adaption of chess, where players receive limited information about the opponent's moves. When training agents to play this game, it is highly useful to be able to evaluate a specific situation (i.e. the received observations at one point in time), and evaluation functions for regular chess are readily available (e.g., from open-source programs such as Stockfish\footnote{\url{https://stockfishchess.org/}}. Thus, computing the average evaluation of all states in an information set is an intuitive approach, but doing so is largely unfeasible due to the information set size. 

\subsubsection{Setup}

For this experiment, training data is created offline for each $k$, thus generating a fixed training set for each setting. Each learner has a fixed budget of 1 million state evaluations that can be arbitrarily distributed among different information sets. Based on the previous results, sensible values of $k$ were chosen as\{1,2,3,5,10,25,50,100,1000\}, thus resulting in datasets of approximately \{1M, 500k, 333k, 200k, 100k, 40k, 20k, 10k, 1k\} examples respectively.\footnote{The exact numbers vary slightly because the information set can consist of fewer states than $k$, thus exhausting it completely.} Importantly, the number of potential public-information states to get evaluated in this experiment is much higher than in the previous Section~\ref{sec:Poker}. For \textit{Poker}, there are only 1326 unique 2-card hands. However, in \textit{RBC}, the number of potential observations which form one information set is estimated to be $10^{139}$ \cite{markowitz2018complexity}, much larger than our training datasets, thus minimising the probability of overlapping training example and increasing the importance of meaningful target value estimations.

\subsubsection{Results}

\begin{figure}[h]
    \centering
    \includegraphics[width = 0.9\linewidth]{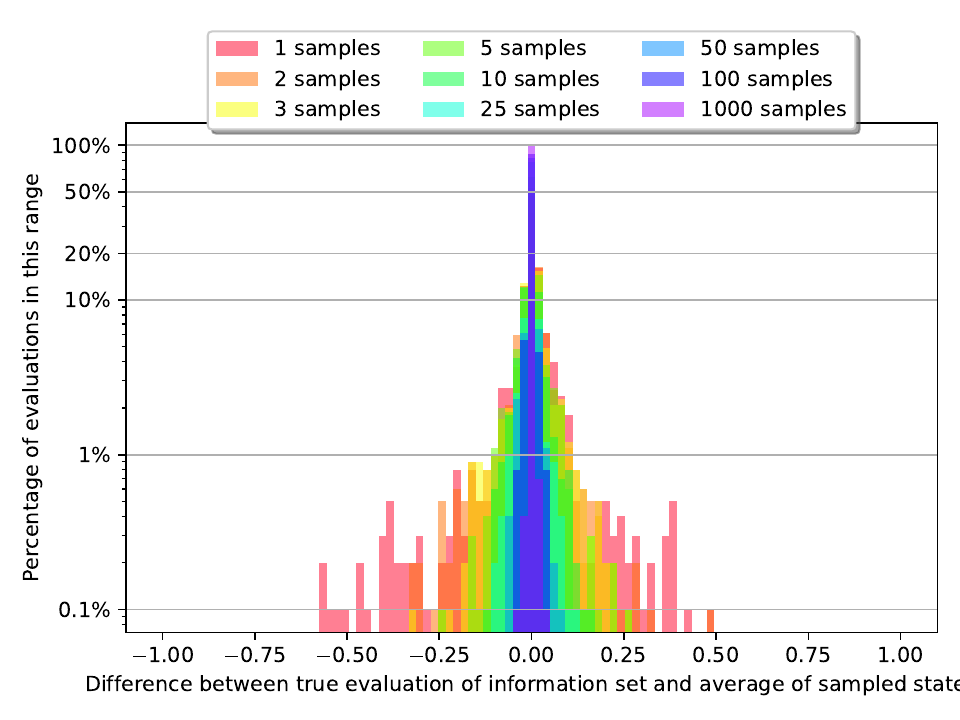}
    \caption{Monte-Carlo estimate of error in evaluating the odds of winning for a given observation. Estimations are computed by averaging over $k$ samples of possible board states, true evaluation is defined as the average over the whole information set. Note that the y-axis is in a logarithmic scale to improve readability.}
    \label{fig:rbc_eval_error_samples}
\end{figure}
\begin{figure}[h]
    \centering
    \includegraphics[width = 0.9\linewidth]{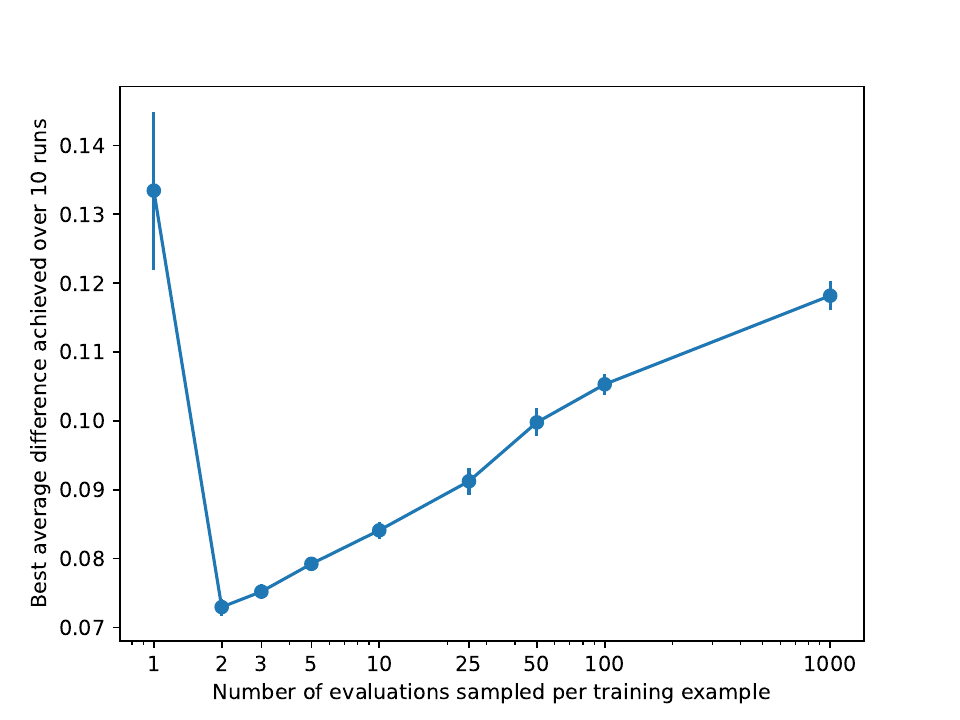}
    \caption{Average lowest received error for training datasets created with a total budget of 1M sampled boards. For each training example, $k$ different boards are sampled from the information set, leading to 1M/$k$ different training examples.}
    \label{fig:rbc_best_version}
\end{figure}

Again approximating the sampling error through Monte-Carlo estimates (Figure~\ref{fig:rbc_eval_error_samples}), we receive the expected result: Sampling only a small number of states can lead to a large discrepancy between the approximation and the ground truth, but we see diminishing returns, such that sampling more than 50 positions only leads to slight improvements in the approximation. We thus expect efficiency increases with repeated sampling, but more than 50 states should lead to meaningfully degrading performance.

\begin{figure}[tb]
    \centering
    \includegraphics[width = 0.9\linewidth]{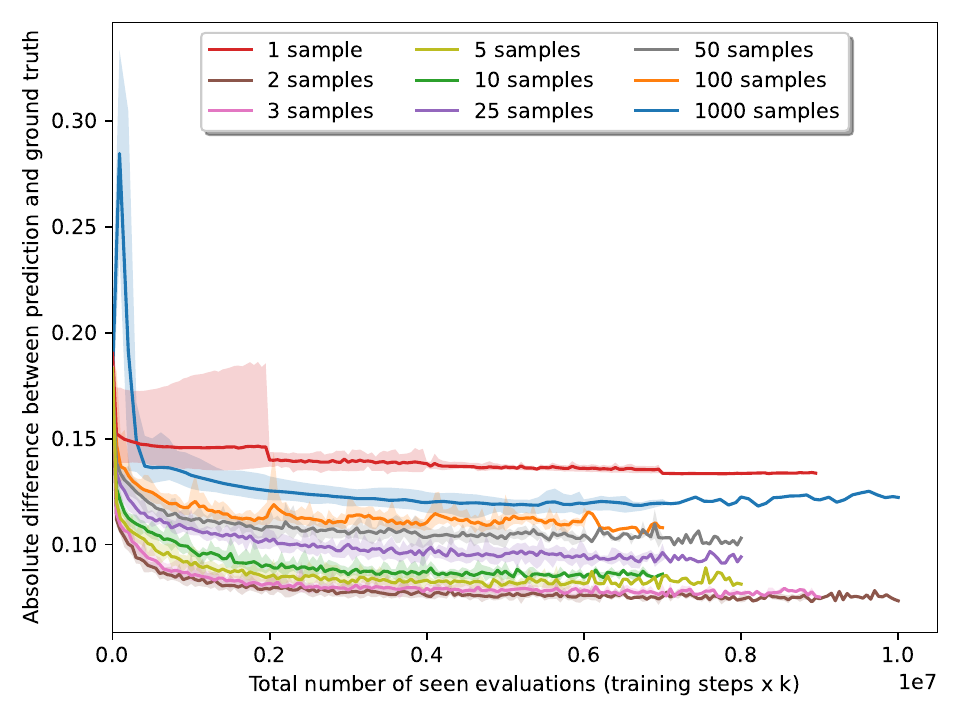}
    \caption{Mean training progress of learning to evaluate imperfect information states with different numbers of boards sampled per training example. The x-axis is scaled by the total number of evaluations seen. Curves vary in length because the Neural Networks are individually trained until no further improvements are seen.}
    \label{fig:rbc_training_curve}
\end{figure}

This observation is confirmed by the results in Figures~\ref{fig:rbc_best_version} and~\ref{fig:rbc_training_curve}. The training curves of Figure~\ref{fig:rbc_training_curve} show that using singular samples results in poor overall accuracy, but vast oversampling leads to too few total training examples. The extreme choices of \textit{k} (1 and 1000) perform poorly, but moderate sampling is useful.

\section{Summary and Conclusion}
\label{sec:summary_conclusion}

With this work, we provided experimental results on the influence of sampling different numbers of states from an information set to learn an evaluation of the whole set. For a given task, a total budget of $N$ evaluations is given, which can be distributed across information sets. Thus, we investigated the trade-off between the overall number of training samples generated and the accuracy of their associated training targets. 

Firstly, the trade-off is influenced by the cost of generating evaluations and the cost of making an update to the learner, so the choice of $k$ needs to be related to their balance. We find that in \textit{Heads Up Poker} and \textit{Reconnaissance Blind Chess} (Sections~\ref{sec:Poker} and \ref{sec:RBC}), multiple evaluations consistently improve the performance and efficiency of the learner. In both domains, sampling two evaluations per training example leads to the overall best results, and only using one sample did not perform well in comparison. As the results for both tasks were similar, we speculate that these findings will translate to more scenarios, but more work is required to validate this. 

\section{Future Work}
\label{sec:future_work}

We see multiple intriguing lines of further work based on these findings. First, we here assumed no agency over the process of sampling from the information sets and no online variation in sample numbers. Removing either of those assumptions will likely lead to better results, and some strategies have previously been outlined by Sheng\cite{sheng2008get} for categorical tasks. In addition, while our general formulation holds for other distributions of states, we used a uniform distribution of states for our experiments. This is a sensible assumption for \textit{Poker}, but information sets do not follow a uniform distribution in \textit{RBC}. Knowledge of this, or even access to a proxy of such a distribution, would lead to more accurate estimations in real-world tasks. Whether a non-uniform distribution changes the best choice of sampled evaluations will be investigated in the future.

\newpage
\bibliographystyle{splncs04}
\bibliography{bib}

\begin{thebibliography}{10}
\providecommand{\url}[1]{\texttt{#1}}
\providecommand{\urlprefix}{URL }
\providecommand{\doi}[1]{https://doi.org/#1}

\bibitem{bertram2022supervised}
Bertram, T., F{\"u}rnkranz, J., M{\"u}ller, M.: Supervised and reinforcement learning from observations in reconnaissance blind chess. In: Proceedings of the 2022 IEEE Conference on Games (CoG). pp. 608--611. IEEE (2022)

\bibitem{bertram2023weighting}
Bertram, T., F{\"u}rnkranz, J., M{\"u}ller, M.: Weighting information sets with {S}iamese neural networks in reconnaissance blind chess. In: Proceedings of the 2023 IEEE Conference on Games (CoG). IEEE (2023)

\bibitem{bluml2023alphaze}
Bl{\"u}ml, J., Czech, J., Kersting, K.: Alphaze**: Alphazero-like baselines for imperfect information games are surprisingly strong. Frontiers in Artificial Intelligence  \textbf{6},  1014561 (2023)

\bibitem{brown2020combining}
Brown, N., Bakhtin, A., Lerer, A., Gong, Q.: Combining deep reinforcement learning and search for imperfect-information games. Advances in Neural Information Processing Systems  \textbf{33},  17057--17069 (2020)

\bibitem{browne2012survey}
Browne, C.B., Powley, E., Whitehouse, D., Lucas, S.M., Cowling, P.I., Rohlfshagen, P., Tavener, S., Perez, D., Samothrakis, S., Colton, S.: A survey of monte carlo tree search methods. IEEE Transactions on Computational Intelligence and {AI} in games  \textbf{4}(1),  1--43 (2012)

\bibitem{furtak2013recursive}
Furtak, T., Buro, M.: Recursive {M}onte {C}arlo search for imperfect information games. In: 2013 IEEE Conference on Computational Inteligence in Games (CIG). pp.~1--8. IEEE (2013)

\bibitem{RBC-tournament-1}
Gardner, R.W., Lowman, C., Richardson, C., Llorens, A.J., Markowitz, J., Drenkow, N., Newman, A., Clark, G., Perrotta, G., Perrotta, R., Highley, T., Shcherbina, V., Bernadoni, W., Jordan, M., Asenov, A.: The first international competition in machine reconnaissance blind chess. In: Escalante, H.J., Hadsell, R. (eds.) Proceedings of the NeurIPS 2019 Competition and Demonstration Track. vol.~123, pp. 121--130. {PMLR}, Vancouver, Canada (2019)

\bibitem{gardner2022machine}
Gardner, R.W., Perrotta, G., Shah, A., Kalyanakrishnan, S., Wang, K.A., Clark, G., Bertram, T., F{\"u}rnkranz, J., M{\"u}ller, M., Garrison, B.P., et~al.: The machine reconnaissance blind chess tournament of {NeurIPS} 2022. In: Proceedings of the NeurIPS 2022 Competitions and Demonstrations Track. pp. 119--132. PMLR (2023)

\bibitem{GIB}
Ginsberg, M.L.: {GIB:} imperfect information in a computationally challenging game. Journal of Artificial Intelligence Research  \textbf{14},  303--358 (2001)

\bibitem{karger2014budget}
Karger, D.R., Oh, S., Shah, D.: Budget-optimal task allocation for reliable crowdsourcing systems. Operations Research  \textbf{62}(1),  1--24 (2014)

\bibitem{khetan2018learning}
Khetan, A., Lipton, Z.C., Anandkumar, A.: Learning from noisy singly-labeled data. In: Proceedings of the 6th International Conference on Learning Representations (ICLR). OpenReview.net, Vancouver, BC, Canada (2018)

\bibitem{Long2010}
Long, J., Sturtevant, N., Buro, M., Furtak, T.: Understanding the success of perfect information {Monte Carlo} sampling in game tree search. In: Burgard, W., Roth, D. (eds.) Proceedings of the 25th AAAI Conference on Artificial Intelligence. pp. 134--140 (2010)

\bibitem{long2010understanding}
Long, J., Sturtevant, N., Buro, M., Furtak, T.: Understanding the success of perfect information {M}onte {C}arlo sampling in game tree search. In: Proceedings of the AAAI Conference on Artificial Intelligence. vol.~24, pp. 134--140 (2010)

\bibitem{markowitz2018complexity}
Markowitz, J., Gardner, R.W., Llorens, A.J.: On the complexity of reconnaissance blind chess. arXiv preprint arXiv:1811.03119  (2018)

\bibitem{RBC-tournament-2}
Perrotta, G., Gardner, R.W., Lowman, C., Taufeeque, M., Tongia, N., Kalyanakrishnan, S., Clark, G., Wang, K., Rothberg, E., Garrison, B.P., Dasgupta, P., Canavan, C., McCabe, L.: The second {NeurIPS} tournament of reconnaissance blind chess. In: Kiela, D., Ciccone, M., Caputo, B. (eds.) Proceedings of the NeurIPS 2021 Competitions and Demonstrations Track. vol.~176, pp. 53--65. {PMLR} (2021)

\bibitem{PlayerOfGames}
Schmid, M., Moravcik, M., Burch, N., Kadlec, R., Davidson, J., Waugh, K., Bard, N., Timbers, F., Lanctot, M., Holland, Z., Davoodi, E., Christianson, A., Bowling, M.: Player of games. arXiv preprint  \textbf{2112.03178} (2021)

\bibitem{settles2012active}
Settles, B.: Active Learning. Synthesis Lectures on Artificial Intelligence and Machine Learning, Morgan {\&} Claypool Publishers (2012)

\bibitem{sheng2008get}
Sheng, V.S., Provost, F., Ipeirotis, P.G.: Get another label? improving data quality and data mining using multiple, noisy labelers. In: Proceedings of the 14th ACM SIGKDD International Conference on Knowledge Discovery and Data Mining. pp. 614--622 (2008)

\bibitem{lig*Sheppard99}
Sheppard, B.: Mastering {Scrabble}. IEEE Intelligent Systems  \textbf{14}(6),  15--16 (November/December 1999)

\bibitem{silver2018general}
Silver, D., Hubert, T., Schrittwieser, J., Antonoglou, I., Lai, M., Guez, A., Lanctot, M., Sifre, L., Kumaran, D., Graepel, T., et~al.: A general reinforcement learning algorithm that masters chess, shogi, and go through self-play. Science  \textbf{362}(6419),  1140--1144 (2018)

\bibitem{snow2008cheap}
Snow, R., O’Connor, B., Jurafsky, D., Ng, A.Y.: Cheap and fast--but is it good? evaluating non-expert annotations for natural language tasks. In: Proceedings of the 2008 Conference on Empirical Methods in Natural Language Processing. pp. 254--263 (2008)

\bibitem{6031993}
Whitehouse, D., Powley, E.J., Cowling, P.I.: Determinization and information set monte carlo tree search for the card game dou di zhu. In: 2011 IEEE Conference on Computational Intelligence and Games (CIG'11). pp. 87--94 (2011)

\bibitem{zinkevich2007regret}
Zinkevich, M., Johanson, M., Bowling, M., Piccione, C.: Regret minimization in games with incomplete information. Advances in Neural Information Processing Systems  \textbf{20} (2007)

\end{thebibliography}

\end{document}